\documentclass[12pt]{article}
\usepackage[paperheight=11in,paperwidth=8.5in,margin=1in]{geometry}
\usepackage{microtype}
\usepackage{booktabs} % for professional tables

\usepackage{titling}
\usepackage{times}
\usepackage{epsfig}
\usepackage{graphicx}
\usepackage{amsmath}
\usepackage{amssymb}
\usepackage{authblk}
\usepackage{float} 
\usepackage{subfigure}
\usepackage{multirow}
\usepackage{enumerate}
\usepackage{breakcites}
\usepackage{hyperref}
\usepackage{titlesec}
\usepackage{titlesec}
\titlespacing*{\section}{0pt}{\baselineskip}{\baselineskip}

\title{\LARGE Computer Vision and Metrics Learning for Hypothesis Testing: An Application of Q-Q Plot for Normality Test}
\author[1]{Ke-Wei Huang}
\author[1]{Mengke Qiao}
\author[1]{Xuanqi Liu}
\author[1]{Siyuan Liu}
\author[1]{Mingxi Dai}
\affil[1]{Department of Information Systems and Analytics,
National University of Singapore, Singapore. Correspondence to: Ke-Wei Huang huangkw@comp.nus.edu.sg}
\date{}
\begin{document}

\maketitle

\begin{abstract}
This paper proposes a new deep-learning method to construct test statistics by computer vision and metrics learning. The application highlighted in this paper is applying computer vision on Q-Q plot to construct a new test statistic for normality test. To the best of our knowledge, there is no similar application documented in the literature. Traditionally, there are two families of approaches for verifying the probability distribution of a random variable. Researchers either subjectively assess the Q-Q plot or objectively use a mathematical formula, such as Kolmogorov-Smirnov test, to formally conduct a normality test. Graphical assessment by human beings is not rigorous whereas normality test statistics may not be accurate enough when the uniformly most powerful test does not exist. It may take tens of years for statistician to develop a new test statistic that is more powerful statistically. Our proposed method integrates four components based on deep learning: an image representation learning component of a Q-Q plot, a dimension reduction component, a metrics learning component that best quantifies the differences between two Q-Q plots for normality test, and a new normality hypothesis testing process. Our experimentation results show that the machine-learning-based test statistics can outperform several widely-used traditional normality tests. This study provides convincing evidence that the proposed method could objectively create a powerful test statistic based on Q-Q plots and this method could be modified to construct many more powerful test statistics for other applications in the future.
\end{abstract}
\section{Introduction}

 Normality assumption is one of the most widely imposed assumptions in many statistical procedures such as t-tests, linear regression analysis, and Analysis of Variance (ANOVA) \cite{razali2011power}. If the assumption of normality is violated, interpretation and inference may not be reliable. Therefore, it is crucial to verify this assumption before conducting further statistical analysis. Traditionally, two common ways to examine the normality assumption are graphical exploration by Q-Q plot (quantile-quantile plot) and formal normality tests.
 
 Q-Q plot is a widely used and effective visualization tool for assessing the empirical probability distribution of a random variable, against any hypothesized theoretical distribution. Q-Q plot compares two probability distributions by plotting theoretical quantile (horizontal axis) against empirical quantile (vertical axis). For example, Q-Q plot is commonly used in linear regression analysis to examine whether regression residuals are normally distributed. If the residuals are normally distributed, the Q-Q plot will exhibit a pattern similar to a 45-degree line. Figure 1(a) illustrates this case. Figure 1(b) illustrates the Q-Q plot of a random variable drawn from Laplace distribution. Throughout this paper, Q-Q plots of a variable drawn from normal distribution are referred to as null hypothesis plots and are abbreviated as H0 plots. On the contrary, plots of random variables drawn from non-normal distribution are referred to as alternative hypothesis plots and are abbreviated as H1 plots.

\begin{figure}[ht]
\centering
\subfigure[Normal]{
\label{Fig.sub.3}
\includegraphics[width=0.4\textwidth]{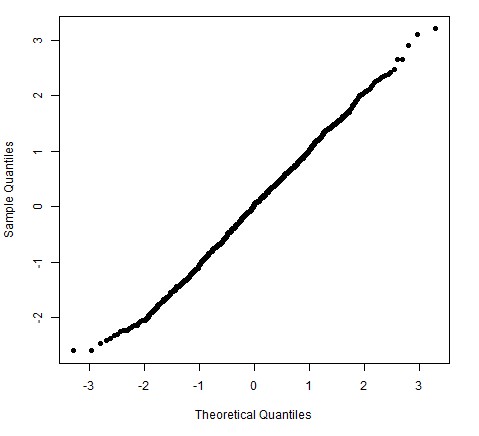}}
\subfigure[Laplace]{\label{Fig.sub.4}

\includegraphics[width=0.4\textwidth]{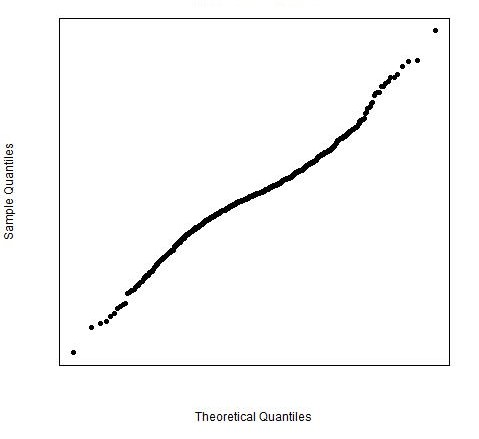}}
\caption{Q-Q Plots for Normal and Laplace Random Variables}

\label{fig:short}
\end{figure}

 Although the graphical method can serve as an effective tool in checking normality, results are still error-prone in that human beings may not be able to make correct and consistent assessment, particularly when the target distribution for testing is very similar to a normal distribution. One motivation of this study is to explore whether this kind of human task can be automated by machine learning methods. 

If computer vision can automate this visualization task, the other research objective is to investigate whether the new AI-based method can outperform existing normality tests developed by statisticians. In other words, can AI be used to construct more powerful test statistics? There are a compelling number of normality tests available in the literature. The most common normality tests provided in statistical software packages include Kolmogorov-Smimov (KS) test, Anderson-Darling (AD) test, Jarque-Bera (JB) test, Glen-Leemis-Barr test (GLB), Gel-Gastwirth test (GG) and Bonett-Seier test (BS). Explanations of these six methods are deferred to Section 2.1. It is well-known in the statistics literature that the uniformly most powerful statistical test may not exist when the alternative hypothesis is multi-dimensional. In other words, there exist room for improvement in statistical power of existing normality tests because there are numerous alternative distributions. Using Q-Q plot could holistically capture richer information than the information used in each one of the traditional statistical tests. For example, when compared with KS or AD tests, Q-Q plot displays very similar information in that it visualizes the distance between an empirical distribution and theoretical distribution. Q-Q plot can also indirectly visualize the dispersion and moments of the sample data, which are the key variables used in deriving the formula of JB test. 

In this paper, we propose a new method, called Deep Normality Test (DNT), to construct test statistics by applying computer vision and metrics learning on Q-Q plots. To construct the test statistics for DNT, we need to first quantify the similarity between two Q-Q plots. Therefore, the proposed method utilizes deep convolution neural network to extract Q-Q plot image representation features, which map Q-Q plots to numerical vectors. After extracting the numerical features, DNT conducts the metric learning to quantify the differences between two Q-Q plots by LMNN metric learning \cite{weinberger2009distance}. Given the optimized similarity metric, we can compute the distance of each H0 plot to the centroid of all H0 plots and this distance metric is also the test statistics of DNT. Similar to traditional normality test, we can computationally derive the distribution of this test statistic by simulating a large number of H0 plots. Next, we can use the 95\% percentile of the distribution as the cutoff value for rejecting H0. In other words, when conducting DNT on a new random variable's Q-Q plot, we calculate its distance metric to the centroid. If the distance is larger than the 95\% cutoff, we reject the null hypothesis and conclude that this new Q-Q plot is not created based on a normal distribution. Otherwise, we cannot reject the null hypothesis that the new Q-Q plot is created based on a random variable drawn from normal distribution.

To evaluate the performance of this new method relative to traditional statistical tests, we simulate datasets with fourteen types of non-normal distributions, including \textit{t} distribution, uniform distribution, beta distribution, Laplace distribution, gamma distribution and chi-square distribution. Our experimentation results show that our proposed approach on average delivers the best performance when compared with 6 traditional tests. Our results may contribute to the literature in the following ways. First, our method demonstrates a completely different avenue for constructing test statistics with superior performance, at least for normality test. Second, the test statistic itself can improve the research quality in many applications due to the importance of normality assumption. Third, our method could be generalized to construct test statistics for other problems. Fourth, our results could also inspire statisticians to derive new formulas for constructing new test statistics after inspecting the properties of the test statistic constructed from machine learning methods. Last but not the least, this study provides a preliminary evidence that human visualization for statistical plots can be automated by machine learning methods.
 
\section{Related Works}
Our research objective is to develop a new method that outperforms existing non-AI based statistical tests. In this section, we will first review the famous traditional normality tests used for comparison in the experimentation section. Next, we will briefly review the literature for image classification and similarity metrics learning by deep learning.

\subsection{Traditional Normality Tests}
In statistics, researchers have developed many normality tests. Generally, normality tests can be classified into several types based on empirical distribution, moments, and other special tests \cite{yap2011comparisons}. Kolmogorov and Smirnov introduced the first empirical distribution based test for normality \cite{kolmogorov1933sulla,smirnov1948table}. This test is a non-parametric test based on the largest vertical difference between the hypothesized and empirical distribution. Another commonly used test based on empirical distribution is Anderson-Darling test \cite{anderson1952asymptotic}. Unlike the KS test which is distribution-free, the AD test makes use of the specific hypothesized distribution and gives more weight to the tails of the distribution. Using order statistics, Glen, Leemis and Barr proposed one test to judge the fit of a distribution to data \cite{963129}. In addition to the above tests, several moment based tests were developed in the literature. Jarque and Bera proposed one test using the sample standardized third and forth moments \cite{jarque1980efficient}. The JB test performs well for distributions with long tail, while its power is poor for distributions with short tails \cite{thadewald2007jarque}. Gel and Gastwirth proposed a robust version of JB test (RJB) by a robust measure of variance \cite{gel2008robust}. Bonett and Seier developed a test of kurtosis that has high uniform power across a very wide range of symmetric non-normal distributions \cite{bonett2002test}.

\subsection{Image Classification by Deep Learning}
Recently, deep convolutional neural networks have led to breakthroughs for image classification. LeCun et al. proposed the first convolutional neural network LeNet-5 with only two convolutional layers and less than one million parameters to classify handwritten digits \cite{lecun1989backpropagation}. Later, Krizhevsky et al. won the ImageNet competition using a network named AlexNet with 60 million parameters \cite{krizhevsky2012imagenet}. More recently, a succeeding winning network architecture is VGG-Net, which consists of 16-19 layers \cite{Simonyan2014}. This neural network exhibit a simple yet effective strategy of constructing very deep networks: stacking building blocks of the same shape. This strategy is inherited by Residual learning network named ResNet \cite{he2016deep}. Unlike VGG-Net that learns an underlying mapping $H(x)$ fitted by stacked layers, ResNet explicitly approximates a residual function $F(x):=H(x)-x$. This residual network is easier to optimize and can surpass the 100-layer barrier. 

\subsection{Similarity Metrics Learning}

Learning a similarity metric between two images is an important research topic with wide range of applications such as image search. Many prior studies have proposed various algorithms to learn such metrics. 

One pioneering algorithm was proposed by \cite{xing2003distance} that aims at finding ``optimal" metrics for clustering algorithms with additional information about relative similarity. Given relative similarity between pairs of points in \(\mathbb{R}^n\), the authors attempt to learn a distance metric from the Mahalanobis distance family, such that points belonging to different clusters can be identified more effectively. More specifically, the distance metric is of the form \( d(x,y)=\parallel x-y\parallel_A=\sqrt{(x-y)^TA(x-y)} \) whereas $A$ are weights for optimization. The relative similarity is characterized by subsets of observations $S$ and $D$, respectively. $S$ is a set of pairs of points known to be similar, and $D$ is a set of dissimilar pairs. The optimization problem is given by
    
\[\min \limits_{A} \sum_{(x_i, x_j \in S)} \parallel x_i-x_j \parallel_A^2\]
   
\[ s.t. \sum_{(x_i,x_j)\in D} \parallel x_i-x_j \parallel_A \geq 1,\]
This convex optimization problem can be efficiently solved by the combination of gradient ascent and iterative projection algorithm over $A$. After the pioneering work of  \cite{xing2003distance}, there exists an explosive number of studies about image metrics learning. Please refer to \cite{yang2006distance} and \cite{kulis2013metric} for detailed review of the literature.

Among the extant metrics learning literature, our algorithm performs the best based on the local metric learning approach pioneered by Large Margin Nearest Neighbors (LMNN)
\cite{weinberger2009distance}. LMNN is one of the most widely-used Mahalanobis distance learning methods and has been the main algorithm of many applications. One of the reasons for its popularity is that the patterns are learned in a local approach: the learned metric will ensure that the \(k\) nearest neighbors (the ``target neighbors") of any training instance belong to the same class while pushing away training instances of different classes (the ``impostors"). Formally, $S$ is defined as a set of pairs of points known to be the same class within the \(k\)-neighborhood of focal point, and $D$ is a set of pairs from different classes within the \(k\)-neighborhood of focal point. The distance is learned using the following convex problem:
 
\[  \min   \sum_{(x_i, x_j \in S)} \parallel x_i-x_j \parallel_A^2+\mu \sum_{ijk} \eta_{ijk}\]
    
\[s.t. \sum_{(x_i,x_k)\in D} \parallel x_i-x_k \parallel_A^2- \sum_{(x_i,x_j)\in S} \parallel x_i-x_j \parallel_A^2 \geq 1-\eta_{ijk},\]

\section{Methods}

In this section, we explain our DNT framework in detail. The proposed framework is illustrated in Figure 2. It consists of four parts: an image representation learning component of a Q-Q plot, a dimension reduction component, a metrics learning component that best quantifies the differences between two Q-Q plots for normality test, and a new normality hypothesis testing process. The first three parts are used to measure the differences between two Q-Q plots. The last part is used to derive the newly constructed test statistic.  
\begin{figure*}[ht]
\centering
\includegraphics[width=14cm]{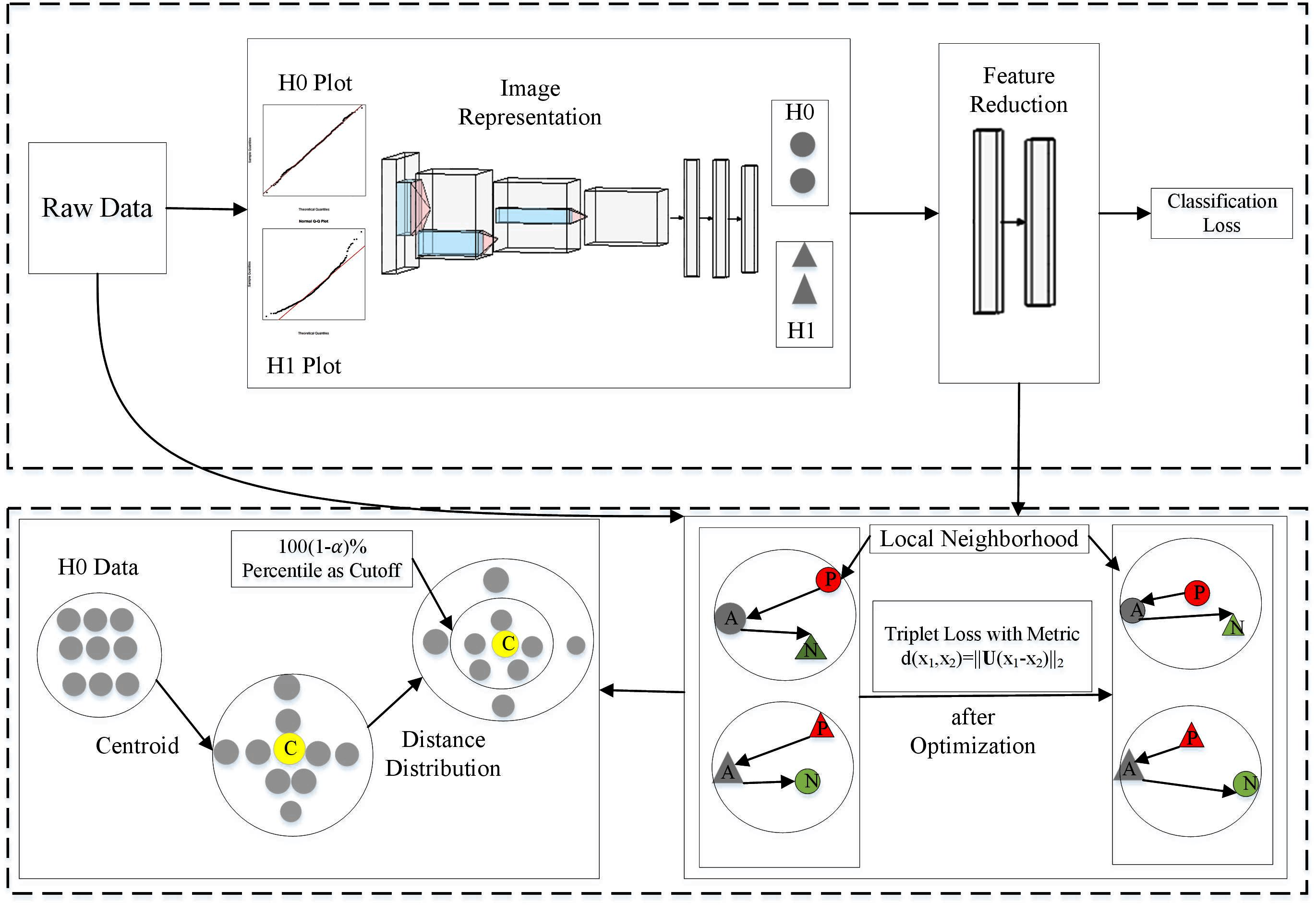}
\caption{Deep Normality Test Framework}
\label{fig:short}
\end{figure*}

\subsection{Design Details}
\subsubsection{Feature Extractor and Dimension Reduction}

Let \(\{(M_1,y_1),...,(M_N,y_N)\} \in R^m \times H \) be the training data with discrete labels \(H=\{0,1\}\) that indicates the plot is H0 or H1 plot. \(M_{i}\) represents the \(i\)-th image in the dataset and \(N\) is the sample size of images in the training dataset. The embedding of the image can be described as a function \(f(M_i)\in R^d\), where \textit{d} \textless \textit{m} is the reduced dimensionality of the Euclidean Space for metrics learning. The deep neural network structure for image feature extraction can be LeNet, AlexNet, VGG, ResNet, or Inception. VGG-19 will be employed as the main model due to its superior performance in our experimentation.

From our experimentation, using the original image embeddings from VGG or other models do not perform well in general because the built-in dimension reduction capability (if exists) of metrics learning algorithms are not effective enough in solving our problem. Adding a dimension reduction or features selection component can improve the overall performance significantly. Conceptually, any dimension reduction methods (e.g., PCA, Kernel-PCA, or SVD) or features selection methods (e.g, Boruta) can be inserted here. Our experimentation results suggest that using XGBoost to select features by importance scores achieve the best performance. The objective function for features selection by XGBoost is by minimizing the classification loss.
\[L_C=-\frac{1}{N}\sum_{i=1}^{N} y_i log p_i +(1-y_i)log(1-p_i).\]

It is worth highlighting that seeding some H1 plots into the training procedure is vital for improving the overall performance of this framework. The reason can be explained by the following example. With only H0 plot, if we try to assess the value of information of one feature, we cannot differentiate whether small variance of that feature is beneficial or harmful because the value of information depends on the mean value of that feature of H0 plots versus mean value of that feature of H1 plots. If the mean value is the same, that feature is useless to differentiate two types of plots. If the mean value is different, then smaller variance helps in differentiating H0 plots from H1 plots. All traditional normality tests seem like using assumptions only under null hypothesis. However, statisticians already hand-pick the ``features" that can best differentiate normal distribution from others. For example, some tests are derived based on the fact that Kurtosis of normal distribution equals to 3, which is a good example that this feature is chosen because ``domain knowledge" suggests that this is unique trait of normal distribution. 

\subsubsection{Triplet Loss for Metric Learning}

After we transform the images into vectors in Euclidean Space, we conduct metric learning by minimizing the triplet loss based on LMNN \cite{weinberger2009distance}. For an image triplet \(\{f(M_i), f(M_i^+), f(M_i^-) \}\in f(M)\), \(f(M_i)\) is the focal point, \((f(M_i), f(M_i^+))\) is termed as a positive pair of Q-Q plots within the same class, and \((f(M_i), f(M_i^-))\) is termed as a negative pair of Q-Q plots that belong to different classes. To construct triplet, for each \(f(M_i)\), we select \(k\) closest positive target neighbors (according to the Euclidean distance) within the same class. Moreover, we select the negative neighbors from different class within the local neighborhood of \(f(M_i)\). We will learn a Mahalanobis metric that best distinguishes H0 plots and H1 plots. The metric can be viewed as a generalization of the Euclidean metric,

\begin{small}
\begin{align*}
 d(f(M_i), f(M_j)) 
&=\sqrt{(f(M_i)-f(M_j))^TM(f(M_i)-f(M_j))}\\
&=\parallel U^T f(M_i)-U^{T} f(M_j)\parallel _2,\\
\end{align*}
\end{small}
where \(M=UU^T\) is a positive semi-defined matrix, and \(U \in R^{d\times d}\) is a transformation matrix. The objective function based on triplet loss is defined as,

\begin{small}
\begin{align*}
&minL =\sum_{(f(M_i), f(M_i^+))}d(f(M_i),f(M_i^+))^2+\\
&\sum_{(f(M_i), f(M_i^-))}[1+d(f(M_i),f(M_i^+))^2-d(f(M_i), f(M_i^-))^2]_+
\end{align*}
\end{small}

During the minimization of the triplet loss, the point of \(f(M_i^+)\) is pulled towards the focal point \(f(M_i)\) and the point of \(f(M_i^-)\) is pushed in the opposite direction with respect to the focal input point. After training, the inputs of the same class are closer to each other than the other class, measured by the new Mahalanobis metric distance.

\subsubsection{Test Statistic Distribution Creation}

After training the first three parts, we can measure the similarity between any two Q-Q plots. The idea of our method is to identify a centroid of all H0 plots. If a new plot is similar enough to the centroid, then it is classified as H0 plot.

Our approach utilizes Monte-Carlo method to derive the test statistic distribution. Specifically, we simulate a large number of H0 plots based on standard normal distribution, \((T_1, T_2,...,T_N)\). Next, each simulated Q-Q plot will be converted into a numerical vector by our trained feature extractor with dimension reduction, \((f(T_1), f(T_2),...,f(T_N))\). Then, the centroid of all H0 numerical vectors is calculated by the simple mean of all vectors of H0 plots, 
\[C=\frac{1}{N}\sum_{i=1}^{N} f(T_{i}) \quad \]
Centroid is constructed by mean value because theoretically it can minimize the sum of Euclidean distances. Empirically, we experimented with other centroids, such as median, theoretically best Q-Q plot, or clustering methods. However, performance of those more complicated centroid is worse than using simple mean as the centroid.

Given the centroid, the distance metric of each Q-Q plot to the centroid is the new test statistic. By simulating a large number of H0 plots, we can construct the probability distribution of the new test statistic. Similar to traditional normality test, we need to decide a cut-off value that accepts  \(100(1-\alpha)\)\% of H0 plots. The typical value of $\alpha$ is 5, which implies the type-I error of our method is always fixed at 5\% by design.

When we apply DNT on a new Q-Q plot, we will calculate the distance to centroid by
\[d(f(M_i), C)=\parallel U^T f(M_i)-U^{T} C\parallel _2^2\]
Finally, the data point will be labeled as normal if the distance is less than the \(100(1-\alpha)\%\) cut-off value. Otherwise, it will be labeled as non-normal. Therefore,
\[H_{0}=\{{M_{i}: \parallel U^Tf(M_{i})-U^TC\parallel^2_2 \leq d_{100(1-\alpha)\%} }\}\]
\[H_{1}=\{{M_{i}:\parallel U^Tf(M_{i})-U^TC\parallel^2_2 > d_{100(1-\alpha)\%}}\}\] 

%-------------------------------------------------------------------------
\section{Experimentation}
Our visualization for the Q-Q plots is very similar to traditional ones, with theoretical quantiles and sample quantiles displayed as the \textit{x}-axis and \textit{y}-axis, respectively. Specifically, Q-Q plot is created by qqnorm() command in R with a red line added by abline() command. We add this red 45-degree line (\(y = x \) line) to highlight the possible deviation from normality. Without the anchoring 45-degree line, both human beings and computer vision may not be able to detect the deviations from perfect normality.

To demonstrate the superior performance of a newly proposed test statistic, researchers generally need to extensively simulate cases under the alternative hypothesis and show that the new method can reject more cases than traditional methods given a fixed Type-I error. In the experimentation, the type I error (significance level) is fixed at 5\% whereas the power of each method will be compared. Following \cite{razali2011power} that evaluates the statistical power of normality tests, we simulate 14 cases of non-normal distributions with sample size 100 to examine the statistical power of our new method. The alternative distributions are seven symmetric distributions: \(t (2)\), \(t (5)\), \(t (10)\), \(t(50)\), \(U (0,1)\),  \(Beta (2,2)\) and Laplace, and seven asymmetric distributions: \(Beta (6,2)\), \(Beta (3,2)\), \(Beta (2,1)\), \(Gamma (1,5)\), \(Gamma (4,5)\), \(\chi^2(4)\) and \(\chi^2(20)\). The detailed specifications of these 14 distributions are provided in Table 1.
\begin{table}[]
\caption{14 cases of non-normal distributions}
\centering
\begin{tabular}{|c|c|c|}
\hline
Cases & Type & Specification \\ \hline
1 & \textit{t} distribution & $e\sim t(2)$ \\ \hline
2 &  \textit{t} distribution & $e\sim t(5)$ \\ \hline
3 &  \textit{t} distribution & $e\sim t(10)$ \\ \hline
4 &  \textit{t} distribution & $e\sim t(50)$ \\ \hline
5 &  Uniform distribution & $e\sim U(0,1)$ \\ \hline
6  & Beta distribution &  $e\sim Beta(2,2)$ \\ \hline
7& Laplace distribution & $e\sim Laplace(0,1)$
 \\ \hline
8 &  Beta distribution & $e\sim Beta(6,2)$ \\ \hline
9 &  Beta distribution & $e\sim Beta(3,2)$ \\ \hline
10 & Beta distribution & $e\sim Beta(2,1)$ \\ \hline
11 &  Gamma distribution & $e\sim Gamma(1,5)$ \\ \hline
12 & Gamma distribution & $e\sim Gamma(4,5)$ \\ \hline
13 & \begin{tabular}[c]{@{}c@{}} Chi-square\\   distribution\end{tabular} &   $e\sim \chi ^{2}(4)$ \\ \hline
14 & \begin{tabular}[c]{@{}c@{}} Chi-square\\   distribution\end{tabular} &  $e\sim \chi ^{2}(20)$ \\ \hline
\end{tabular}
\end{table}
%\begin{figure}[]
%\centering
%subfigure[]{
%label{Fig.sub.9}
%includegraphics[width=0.2\textwidth]{std.png}
%includegraphics[width=0.2\textwidth]{t.png}
%
%caption{Q-Q plots for normal and \(t(2)\) distributions}
%label{Fig.lable}
%end{figure}

Our training dataset includes 1000 Q-Q plots of \(t(50)\) as H1 plots. According to the statistics literature, for \(t\) distribution, when the degree of freedom increases to infinity, \(t\) distribution converges to standard normal distribution. Therefore, \(t(50)\) is most similar to normal distribution out of 14 test cases. This also implies that we use minimal information about H1 plots during the training procedure while the performance of DNT can still outperform traditional tests.

For training the image features extractor, in addition to 1000 H1 plots, we simulate 50,000 H0 plots and we select only 1\% H0 plots (500 H0 plots) closest to the hypothetical centroid. We find that using H0 plots closer to centroid outperforms using all H0 plots with LMNN metrics learning. The sample size 500 H0 plots and 1000 H1 plots are chosen to be a moderate value to demonstrate the effectiveness of DNT. When the size is too small, the performance of DNT cannot outperform traditional tests. When the sample size further increases, the performance of DNT can be marginally improved. The ratio between H0 and H1 plots is one hyper-parameter for tuning in DNT.

After the feature extractor training and metric learning, we utilize the same 50,000 H0 plots to derive the 95\% rejection cutoff value. For the 14 cases of H1 plot and H0 plot in the test set, we simulate 1000 plots for each of the 15 cases respectively. All Q-Q plots used in the test set are not used in the training process.

We also conduct two sets of robustness checks to prove the effectiveness of DNT by applying computer vision on Q-Q plot. First, it is possible that directly using ordered raw data as input features into LMNN can also construct a powerful test statistic because we are using all order statistics for metrics learning. Intuitively, it is possible that order statistics contain more information than Q-Q plot image features. Therefore, we include ordered raw data as the input features for LMNN without using Q-Q plot's features. Second, some researchers may argue that using traditional image similarity metrics may outperform deep learning models, such as VGG. Therefore, we also investigated using metrics from image quality assessment (IQA) literature, which documents several hand-crafted image similarity measures to assess the image quality. Specifically, PSNR is one of the most extensively used measures due to its simplicity. However, PSNR cannot capture the human perception of image fidelity and quality \cite{Wang2009}. Therefore, some similarity measures following top-down strategy have been proposed \cite{Lin2011,Wang2004}. Specifically, those measures first extract different image features to calculate a similarity map and then summarize the values of the similarity map into a single similarity score. For example, the classic structural similarity index (SSIM) employ the luminance, contrast and structural information to construct a similarity map. SSIM then uses averaging pooling to calculate the final similarity score \cite{Wang2004}. Feature similarity index (FSIM) uses phase congruency and gradient magnitude features to constitute the similarity map \cite{Zhang2011}. Visual Saliency Induced Index (VSI) uses saliency-based features and gradient magnitude for image evaluation \cite{Zhang2014}. Mean deviation similarity index (MDSI) combines gradient similarity, chromaticity similarity, and deviation pooling to enhance the image evaluation \cite{Nafchi2016}. Structural contrast-quality index (SCQI) utilizes Structural Contrast Index to capture local visual quality perception in terms of different image texture features \cite{Bae2016}. These six image similarity measures will be experimented in this paper as another set of benchmark cases. 

\section{Results}
% Please add lts}the following required packages to your document preambl:
% \usepackage{multirow}
% \usepackage{multirow}

\begin{table}[]
\caption{ DNT v.s. traditional normality test results}
\centering
\begin{tabular}{|c|c|c|c|c|c|c|}
\hline
Cases & VGG & AD    & KS    & JB    & GLB   & GG    \\ \hline
1     & 0.964 & 0.978 & 0.955 & 0.984 & 0.978 & 0.991 \\ \hline
2     & 0.535 & 0.464 & 0.317 & 0.643 & 0.466 & 0.671  \\ \hline
3     & 0.208 & 0.154 & 0.101 & 0.298 & 0.15  & 0.302 \\ \hline
4     & 0.071 & 0.057 & 0.053 & 0.091 & 0.056 & 0.098 \\ \hline
5     & 0.999 & 0.948 & 0.577 & 0.711 & 0.948 & 0.011 \\ \hline
6     & 0.567 & 0.316 & 0.145 & 0.046 & 0.318 & 0     \\ \hline
7     & 0.693 & 0.843 & 0.697 & 0.807 & 0.849 & 0.897 \\ \hline
8     & 0.85  & 0.84  & 0.624 & 0.724 & 0.818 & 0.565 \\ \hline
9     & 0.413 & 0.387 & 0.234 & 0.102 & 0.379 & 0.021 \\ \hline
10    & 0.987 & 0.98  & 0.825 & 0.84  & 0.977 & 0.386 \\ \hline
11    & 1     & 1     & 1     & 1     & 1     & 0.999 \\ \hline
12    & 0.869 & 0.888 & 0.703 & 0.891 & 0.88  & 0.829 \\ \hline
13    & 0.994 & 0.998 & 0.939 & 0.997 & 0.998 & 0.979 \\ \hline
14    & 0.46  & 0.504 & 0.355 & 0.537 & 0.495 & 0.484 \\ \hline
15    & 0.052 & 0.057 & 0.051 & 0.048 & 0.06  & 0.048 \\ \hline
Mean  & \textbf{0.686} & 0.668 & 0.538 & 0.619 & 0.665 & 0.517 \\ \hline
\end{tabular}
\end{table}

Table 2 reports the statistical power (rejection) of DNT and traditional normality tests for the 14 cases of H1 plots described in Table 1. The larger the value, the better the statistical test is. Case 15 is the results of H0 plot and it should be close to 5\% in theory. The row ``mean" reports the average power across all the H1 cases. The column ``VGG" is the proposed DNT test based on VGG-19 feature extractor. The other 5 columns report the statistical power of traditional normality tests described in Section 2.1, including Kolmogorov-Smirnov test (KS), Anderson-Darling test (AD), Jarque-Bera test (JB), Glen-Leemis-Barr test (GLB), and Gel-Gastwirth test (GG). Bonett-Seier test (BS) is omitted due to page limit and its average power is only 0.443, which is much worse than the other 6 columns reported in Table 2. Comparing our method and 6 traditional tests, we can find that the average performance of DNT can outperform all 6 traditional tests. It can be verified that this superior performance persists even we exclude results of Case 4 \(t(50)\). DNT also performs the best in 6 out of 14 cases whereas other tests can be the best-performing case in few cases.

Table 3 reports the results by other benchmark cases. The column ``ResNet" refers to the results of using ResNet, not VGG-19, as the image feature extractor. Interestingly, using different CNN leads to very different DNT performance in our experimentation. The column ``Raw" refers to the results from using ordered raw data as features. The  other 6 columns refer to the results obtained after we use the traditional image similarity metrics. We can reach the following conclusions. First, our computer vision method by using VGG as network structure on average can outperform all benchmark cases, achieving the highest overall power of 0.686. Second, among all the traditional image similarity metrics, MDSI performs the best, achieving the overall power of 0.677. However, MDSI still performs worse than VGG, which indicates the effectiveness of the deep learning network for processing Q-Q plots. Third, using raw data as feature performs worse than our method and MDSI. This result demonstrates the usefulness of applying computer vision method on Q-Q plots to extract useful information. 

In our VGG model, our experimentation shows that the best value of $k$ of LMNN is 25 and the optimal number of features size is around 100. In Table 4, we also report the test results of different number of features after dimension reduction, and different number of neighbors in LMNN. Table 4 suggests that results are not that sensitive to the number of neighbors but are sensitive to the number of features for metrics learning. Due to the page limit, we omit the experimentation results of other models.

\begin{table*}[]
\caption{Comparison results between DNT and benchmark cases}
\centering
\begin{tabular}{|c|c|c|c|c|c|c|c|c|c|}
\hline
Cases & VGG   & ResNet & Raw   & FSIM  & VSI   & SCQI  & PSNR  & SSIM  & MDSI  \\ \hline
1     & 0.964 & 0.933  & 0.98  & 0.427 & 0.926 & 0.945 & 0.31  & 0.37  & 0.973 \\ \hline
2     & 0.535 & 0.417  & 0.5   & 0.363 & 0.418 & 0.458 & 0.186 & 0.117 & 0.475 \\ \hline
3     & 0.208 & 0.147  & 0.164 & 0.179 & 0.182 & 0.212 & 0.103 & 0.064 & 0.172 \\ \hline
4     & 0.071 & 0.067  & 0.059 & 0.077 & 0.08  & 0.084 & 0.051 & 0.052 & 0.061 \\ \hline
5     & 0.999 & 0.635  & 0.931 & 0.889 & 0.811 & 0.576 & 0.876 & 0.996 & 0.984 \\ \hline
6     & 0.567 & 0.236  & 0.295 & 0.028 & 0.14  & 0.004 & 0.091 & 0.427 & 0.278 \\ \hline
7     & 0.693 & 0.507  & 0.86  & 0.585 & 0.535 & 0.463 & 0.19  & 0.44  & 0.747 \\ \hline
8     & 0.85  & 0.713  & 0.798 & 0.282 & 0.928 & 0.846 & 0.956 & 0.873 & 0.942 \\ \hline
9     & 0.413 & 0.299  & 0.353 & 0.07  & 0.365 & 0.109 & 0.391 & 0.556 & 0.408 \\ \hline
10    & 0.987 & 0.841  & 0.967 & 0.293 & 0.979 & 0.958 & 0.987 & 0.999 & 0.998 \\ \hline
11    & 1     & 0.995  & 1     & 0     & 1     & 1     & 0.973 & 0.821 & 1     \\ \hline
12    & 0.869 & 0.707  & 0.882 & 0.112 & 0.947 & 0.885 & 0.945 & 0.752 & 0.937 \\ \hline
13    & 0.994 & 0.947  & 0.998 & 0.03  & 1     & 0.997 & 0.991 & 0.846 & 1     \\ \hline
14    & 0.46  & 0.328  & 0.492 & 0.057 & 0.576 & 0.474 & 0.584 & 0.428 & 0.505 \\ \hline
15    & 0.052 & 0.033  & 0.061 & 0.061 & 0.054 & 0.065 & 0.049 & 0.048 & 0.053 \\ \hline
Mean  & \textbf{0.686} & 0.555  & 0.663 & 0.242 & 0.635 & 0.572 & 0.545 & 0.553 & 0.677 \\ \hline
\end{tabular}
\end{table*}

% \usepackage{multirow}

\begin{table}[]
\caption{DNT results for different feature sizes and number of neighbors}
\centering
\begin{footnotesize}
\begin{tabular}{|c|c|c|c|c|c|c|c|c|c|c|}
\hline
\multirow{2}{*}{Cases} & \multicolumn{10}{c|}{(Number of Neighbors, Feature Size)}                                                 \\ \cline{2-11} 
                       & (25,100) & (25,50) & (25,75) & (25,125) & (25,150) & (15,100) & (20,100) & (30,100) & (35,100) & (40,100) \\ \hline
1                      & 0.964    & 0.976   & 0.965   & 0.957    & 0.958    & 0.954    & 0.958    & 0.959    & 0.967    & 0.968    \\ \hline
2                      & 0.535    & 0.604   & 0.553   & 0.514    & 0.512    & 0.506    & 0.52     & 0.522    & 0.548    & 0.554    \\ \hline
3                      & 0.208    & 0.242   & 0.221   & 0.193    & 0.209    & 0.198    & 0.205    & 0.211    & 0.218    & 0.219    \\ \hline
4                      & 0.071    & 0.074   & 0.07    & 0.065    & 0.063    & 0.061    & 0.069    & 0.067    & 0.072    & 0.075    \\ \hline
5                      & 0.999    & 0.998   & 0.998   & 0.999    & 0.994    & 0.999    & 0.999    & 0.999    & 0.999    & 0.998    \\ \hline
6                      & 0.567    & 0.554   & 0.555   & 0.579    & 0.519    & 0.612    & 0.572    & 0.585    & 0.546    & 0.532    \\ \hline
7                      & 0.693    & 0.734   & 0.687   & 0.658    & 0.65     & 0.648    & 0.672    & 0.67     & 0.691    & 0.696    \\ \hline
8                      & 0.85     & 0.379   & 0.779   & 0.878    & 0.894    & 0.853    & 0.845    & 0.851    & 0.852    & 0.85     \\ \hline
9                      & 0.413    & 0.298   & 0.343   & 0.415    & 0.428    & 0.447    & 0.416    & 0.425    & 0.395    & 0.394    \\ \hline
10                     & 0.987    & 0.912   & 0.984   & 0.991    & 0.995    & 0.988    & 0.986    & 0.984    & 0.985    & 0.982    \\ \hline
11                     & 1        & 0.998   & 0.998   & 1        & 1        & 1        & 1        & 1        & 1        & 1        \\ \hline
12                     & 0.869    & 0.699   & 0.722   & 0.803    & 0.854    & 0.865    & 0.871    & 0.824    & 0.854    & 0.837    \\ \hline
13                     & 0.994    & 0.908   & 0.947   & 0.986    & 0.992    & 0.996    & 0.996    & 0.988    & 0.991    & 0.988    \\ \hline
14                     & 0.46     & 0.39    & 0.343   & 0.402    & 0.449    & 0.445    & 0.453    & 0.38     & 0.434    & 0.423    \\ \hline
15                     & 0.052    & 0.046   & 0.039   & 0.053    & 0.057    & 0.049    & 0.05     & 0.047    & 0.048    & 0.05     \\ \hline
Mean                   & 0.686    & 0.626   & 0.655   & 0.674    & 0.680    & 0.684    & 0.683    & 0.676    & 0.682    & 0.680    \\ \hline
\end{tabular}
\end{footnotesize}
\end{table}

\section{Conclusion and Future Works}
In this study, we propose a new learning method to construct test statistics for normality test based on Q-Q plots. Our methods are built upon the literature about deep-learning based image classification and metrics learning. Our experimentation results show that applying computer vision and metric learning on Q-Q plots delivers superior performance over six widely-used traditional methods. 

This study provides preliminary yet convincing evidence that computer vision on statistical plots can be utilized to construct powerful test statistics. Our method could be generalized to any one- or two-sample distributional test since Q-Q plot can be applied to any one-sample or two-sample distributional test. At the same time, it is convincing that better formatting of Q-Q plot or using multiple statistical plots simultaneously could further improve the performance of the proposed AI-based method for normality test. Furthermore, it seems convincing that our method could be generalized to construct test statistics for other famous hypothesis testing in statistics based on other statistical visualizations, such as residual plots for linear regression analysis or auto-correlation and partial auto-correlation plots for time series analysis. 

\bibliographystyle{acm}
\bibliography{cv}

\end{document}